\documentclass{article}
\usepackage{spconf,amsmath,graphicx}
\usepackage{amssymb}
\usepackage{latexsym}
\usepackage{exscale}
\usepackage{fontenc}
\usepackage{graphicx}
\usepackage{epsfig}
\usepackage{nicefrac}
\usepackage{mathptmx}
\usepackage{hyperref}

\title{Efficient Matrix Completion with Gaussian Models}

\twoauthors
 {Flavien L\'eger}
    {CMLA, ENS Cachan, 61 av. du\\
     Pr\'esident Wilson, Cachan 94235, France}
      {Guoshen Yu, Guillermo Sapiro}
  {ECE, University of Minnesota,\\
   Minneapolis, 55455, MN, USA}

\begin{document}
\newcommand{\ba}{\mathbf{a}}
\newcommand{\bbf}{\mathbf{f}}
\newcommand{\bh}{\mathbf{h}}
\newcommand{\by}{\mathbf{y}}
\newcommand{\bw}{\mathbf{w}}
\newcommand{\bz}{\mathbf{z}}

\newcommand{\bB}{\mathbf{B}}
\newcommand{\bD}{\mathbf{D}}
\newcommand{\bF}{\mathbf{F}}
\newcommand{\bH}{\mathbf{H}}
\newcommand{\bI}{\mathbf{I}}
\newcommand{\bM}{\mathbf{M}}
\newcommand{\bS}{\mathbf{S}}
\newcommand{\bT}{\mathbf{T}}
\newcommand{\bU}{\mathbf{U}}
\newcommand{\bW}{\mathbf{W}}
\newcommand{\bY}{\mathbf{Y}}

\maketitle

\begin{abstract}
A general framework based on Gaussian models and a MAP-EM algorithm is introduced in this paper  for solving matrix/table completion problems. The numerical experiments with the standard and challenging movie ratings data show that the proposed approach, based on 
probably one of the simplest probabilistic models, leads to the results in the same ballpark as the state-of-the-art, at a lower computational cost. 
\end{abstract}

 
\section{Introduction}

Matrix completion amounts to estimate all the entries in a matrix 
$\bF \in \mathbb{R}^{M \times N}$ from a partial and potentially noisy observation 
\begin{equation}
\label{eqn:matrix:completion}
\bY = \bH \bullet \bF + \bW,
\end{equation}
where $\bH \in  \mathbb{R}^{M \times N}$ is a binary matrix with 1/0 entries masking a portion of $\bF$ through the element-wise multiplication $\bullet$, and $\bW \in \mathbb{R}^{M \times N}$ an additive noise. With matrix rows, columns, and entry values assigned to various attributes, matrix completion may have numerous applications. For example, when the rows and the columns of $\bF$ are attributed to users and items such as movies and books, and an entry at position $(m,n)$ records a score given by the $m$-th user to the $n$-th item, matrix completion predicts users' scores on the items they have not yet rated, based on the available scores recorded in $\bY$, so that personalized item recommendation system becomes possible. This is a classical problem in collaborative filtering.

To solve an ill-posed matrix completion problem, one must rely on some prior information, or in other words, some data model. The most popular family of approaches in the literature assumes that the matrix $\bF$ follows approximately a low rank model, and calculates the matrix completion with a matrix factorization~\cite{decoste2006collaborative, rennie2005fast}. Theoretical results regarding the completion of low-rank matrices have been recently obtained as well, e.g., \cite{candes2009exact,recht2009necessary} and references therein. More elaborative probabilistic models and some refinement have been further studied on top of matrix factorization, leading to state-of-the-art results~\cite{lawrence2009non, mackey2010, zhou10nonparametric} . 
 
In image processing, assuming that local image patches follow Gaussian mixture models (GMM), Yu, Sapiro and Mallat have recently reported excellent results in a number of inverse problems~\cite{yu2010PLE}. In particular, for inpainting, which is an analogue to matrix completion where the data is an image, the maximum a posteriori expectation-maximization (MAP-EM) algorithm for solving the GMM leads to state-of-the-art results, with a computational complexity considerably reduced with respect to the previous state-of-the-art approaches based on sparse models~\cite{mallat2008wts}, (which are analogous to the low-rank model assumption for matrices). 
 
In this paper, we investigate Gaussian modeling (a particular case of GMM with only a single Gaussian distribution) for matrix completion. Subparts of the matrix, typically rows or columns, are regarded as a collection of signals that are assumed to follow a Gaussian distribution. An efficient MAP-EM algorithm is introduced to estimate both the Gaussian parameters (mean and covariance) and the signals. We show through numerical experiments that the fast MAP-EM algorithm, based on the Gaussian model which is the simplest probabilistic model one can imagine, leads to results in the same ballpark as the state-of-the-art in movie rating prediction, at significantly lower computational cost. Recent theoretical results~\cite{yu2010scs} further support the consideration of Gaussian models for the recovery of missing data.

Section~\ref{sec:model:algo} introduces the Gaussian model and the MAP-EM algorithm. After presenting the numerical experiments in Section~\ref{sec:numeric}, Section~\ref{sec:conclusion} concludes with some discussions. 
 
\section{Model and Algorithm}
\label{sec:model:algo}

\subsection{Gaussian Model}
\label{subsec:gaussian:model}

Similar to the local patch decomposition often applied in image processing~\cite{yu2010PLE}, let us consider each subpart of the matrix $\bF \in \mathbb{R}^{M \times N}$, the $i$-th row $\bbf_i \in \mathbb{R}^N$ for example, as a signal. Let $\bh_i \in \mathbb{R}^N$ denote the $i$-th row in the binary matrix $\bH$, 
and let $N_i = |\{j|\bh_i(j) \neq 0, 1 \le j \le N\}|$ be the number of non-zero entries in $\bh_i$. Let $\bU_i \in \mathbb{R}^{N_i \times N}$ denote a masking operator which maps from $\mathbb{R}^{N}$ to $\mathbb{R}^{N_i}$ extracting entries of $\bbf_i$ corresponding to the non-zero entries of $\bh_i$, i.e., all but the $(\mathrm{Idx}(\bh_i, k))$-th the entries in the $k$-th row of $\bU_i$ are zero, with $(\mathrm{Idx}(\bh_i, k))$ the index of the $k$-th non-zero entry in $\bh_i$, $1 \leq k \leq N_i$. Let $\by_i \in \mathbb{R}^{N_i}$ and $\bw_i \in \mathbb{R}^{N_i}$ be respectively the \textit{sub-vector} of the $i$-th row of $\bY$ and $\bW$, where the entries of $\bh_i$ are non-zero. With this notation, we can rewrite~\eqref{eqn:matrix:completion} in a more general linear model
\begin{equation}
\label{eqn:problem:vector}
\by_i = \bU_i \bbf_i + \bw_i,
\end{equation}
for all the signals $\bbf_i$, $1 \leq i \leq M$.~\footnote{Writing $\by_i$ in the reduced dimension $N_i$ leads to a calculation in dimension $N_i$ instead of $N$, which is considerably faster if $N_i \ll N$.} Note that $\bbf_i$ can also be columns, or 2D sub-matrices of $\bF$ rendered in 1D. 

The Gaussian model assumes that each signal $\bbf_i$ is drawn from a Gaussian distribution, with a probability density function
\begin{equation}
\label{eqn:gaussian:model}
p(\bbf_i) = \frac{1}{(2\pi)^{N/2} |\Sigma|^{1/2}} \exp\left({-\frac{1}{2} (\bbf_i - \mu)^T \Sigma^{-1} (\bbf_i - \mu)}\right),
\end{equation}
where $\Sigma$ and $\mu$ are the unknown covariance and mean parameters of the Gaussian distribution. The noise $\bw_i$ is 
assumed to follow a Gaussian distribution with zero mean and covariance $\Sigma_{\bw}$, here assumed to be known (or calibrated). 

Estimating the matrix $\bF$ from the partial observation $\bY$ can thus be casted into the following problem:
\begin{itemize}
\item Estimate the Gaussian parameters $(\mu, \Sigma)$, from the observation $\{\by_i\}_{1 \leq i \leq M}$.
\item Estimate $\bbf_i$ from $\by_i$,  $1 \leq i \leq M$, using the Gaussian distribution $\mathcal{N}(\mu, \Sigma)$.
\end{itemize}

Since this is a non-convex problem, we present an efficient maximum a posteriori expectation-maximization (MAP-EM) algorithm that calculates a local-minimum solution~\cite{allassonniere2007towards, yu2010PLE}. 

\subsection{Algorithm}

Following a simple initialization, addressed in Section~\ref{subsec:initialization}, the MAP-EM algorithm is an iterative procedure that alternates between two steps:

\begin{enumerate}
\item E-step: signal estimation. Assuming the estimates $(\tilde{\mu},\tilde{\Sigma})$ are known (following the previous M-step), for each $i$ one computes the maximum a posteriori (MAP) estimate $\tilde{\bbf}_i$ of $\bbf_i$.

\item M-step: model estimation. Assuming the signal estimates $\tilde{\bbf}_i$, $\forall i$, are known (following the previous E-step), one estimates (updates) the Gaussian model parameters $(\tilde{\mu}, \tilde{\Sigma})$. 
\end{enumerate}

\subsubsection{E-step: Signal Estimation}

It is well known that under the Gaussian models assumed in Section~\ref{subsec:gaussian:model}, the MAP estimate that maximizes the log a-posteriori probability
{\small
\begin{eqnarray}
\label{eqn:MAP}
\tilde{\bbf}_{i} 
& = & \arg\max_{\bbf_i} \log p(\bbf_i | \by_i, \tilde{\mu}, \tilde{\Sigma}) \nonumber \\
& =  &\arg\max_{\bbf_i} \left(\log p(\by_i|\bbf_i) + \log p(\bbf_i | \tilde{\mu}, \tilde{\Sigma}) \right) \nonumber  \\
& = & \arg\min_{\bbf_i} \left( (\bU_i \bbf_i - \by_i)^T \Sigma_\bw^{-1} (\bU_i \bbf_i - \by_i) + (\bbf _i- \tilde{\mu})^T \tilde{\Sigma}^{-1}  (\bbf_i - \tilde{\mu})\right) \nonumber  \\
& = & \tilde{\mu} + \tilde{\Sigma} \bU_i^T (\bU_i \tilde{\Sigma} \bU_i^T + \Sigma_\bw)^{-1} (\by_i - \bU_i \tilde{\mu}),
\end{eqnarray}}is a linear estimator and is optimal in the sense that it minimizes the mean square error (MSE), i.e., 
$
E_{\bbf_i, \bw_i}  [\|\bbf_i - \tilde{\bbf}_{i}\|_2^2]  =  \min_{g}E_{\bbf_i, \bw_i}  [\|\bbf _i- g({\by}_i)\|_2^2], \nonumber 
$
as well as the mean absolute error (MAE), i.e., 
$
 E_{\bbf_i, \bw_i}  [\|\bbf_i - \tilde{\bbf}_{i}\|_1]  =  \min_{g} E_{\bbf_i, \bw_i}  [\|\bbf_i - g({\by}_i)\|_1], \nonumber
$
where $g$ is any mapping from $\mathbb{R}^{N} \rightarrow \mathbb{R}^{N_i}$~\cite{kay1998fundamentals}. The second equality of~\eqref{eqn:MAP} follows from the Bayes rule, the third follows from the Gaussian models $\bbf_i \sim \mathcal{N}({\mu}, {\Sigma})$ and $\bw_i \sim  \mathcal{N}(\mathbf{0}, \Sigma_\bw)$, and the last is obtained by deriving the third line with respect to $\bbf_i$. 

The close-form MAP estimate~\eqref{eqn:MAP} can be calculated fast. Observe that $\bU_i \in \mathbb{R}^{N_i \times N}$ is a sparse extraction matrix, each row containing only one non-zero entry with value $1$, whose index corresponds to the non-zero entry in $\bh_i$. 
Therefore, the multiplication operations that involve $\bU_i$ or $\bU_i^T$ can be realized by extracting the appropriate rows or columns at zero computational cost. The complexity of~\eqref{eqn:MAP} is therefore dominated by the matrix inversion. As $\bU_i \tilde{\Sigma} \bU_i^T + \Sigma_\bw$ is positive-definite, $(\bU_i \tilde{\Sigma} \bU_i^T + \Sigma_\bw)^{-1}$ can be implemented with $N_i^3/3$ flops through a Cholesky factorization~\cite{boyd2004convex}. 


In a typical case where $\bbf_i$ is the $i$-th row of the matrix $\bF$, $1 \leq i \leq M$, to estimate $\bF$ the total complexity of the E-step is therefore dominated by $\sum_{i=1}^M N_i^3/3$ flops. For typical rating prediction datasets that are highly sparse, among a large number of items $N$, most users have rated more or less only a small number $N_i \approx N/C$ of items, where $C$ is large. The total complexity of the E-step is thus dominated by $\frac{1}{3 C^3}MN^3$ flops.

\subsubsection{M-step: Model Estimation}
The parameters of the model are estimated/updated with the maximum likelihood estimate,
\begin{equation}
\label{eqn:ML:gaussian}
(\tilde{\mu}, \tilde{\Sigma}) = \arg \max_{\mu, \Sigma} \log p(\{\tilde{\bbf}_i\}_{1 \leq i \leq M}|\mu, \Sigma). 
\end{equation}
With the Gaussian model $\bbf_i \sim \mathcal{N}({\mu}, {\Sigma})$, it is well known that the resulting estimate is the empirical one,
\begin{equation}
\label{eqn:ML:covariacne}
\tilde{\mu} = \frac{1}{M}\sum_{i=1}^M \tilde{\bbf}_i~~\textrm{and}~~
\tilde{\Sigma} = \frac{1}{M} \sum_{i=1}^M (\tilde{\bbf}_i  - \tilde{\mu}) (\tilde{\bbf}_i  - \tilde{\mu})^T. 
\end{equation}

The empirical covariance estimate may be improved through regularization when there is lack of data (let us take an example of standard rating prediction, where there are $N \sim 10^3$ items and $M \sim 10^4$ users, the dimension of the covariance matrix ${\Sigma}_k $ is $N \times N \sim 10^6$). A simple and standard eigenvalue-based regularization is used here,
\begin{equation}
\label{eqn:cov:regularization}
\tilde{\Sigma} \leftarrow \tilde{\Sigma} + \varepsilon Id,
\end{equation}
where $\epsilon$ is a small constant. The regularization also guarantees that the estimate $\tilde{\Sigma}$ of the covariance matrix is full-rank, so that ~\eqref{eqn:MAP} is always well defined. 

To estimate $\bF$, the computational complexity of the M-step is dominated by the calculation of the empirical covariance estimate requiring $MN^2$ flops, which is negligible with respect to the E-step. 

As the MAP-EM algorithm iterates, the MAP probability of the observed signals $p(\tilde{\bbf}_i | \by_i, \tilde{\mu}, \tilde{\Sigma}) 
$ increases. This can be observed by interpreting the E- and M-steps as a coordinate descent optimization~\cite{hathaway1986another}.  In the experiments, the algorithm converges within 10 iterations. 

\subsubsection{Initialization}
\label{subsec:initialization}
The MAP-EM algorithm is initialized with an initial guess of $\bbf_i$, $\forall i$. The experiments show that the result is insensitive to the initialization for movie rating prediction. In the numerical experiments, all the unknown entries are initialized to 3 for datasets containing ratings ranging from 1 to 5 or 6. 

\subsubsection{Computational and Memory Requirements}
In a typical case where the matrix row decomposition in~\eqref{eqn:problem:vector} is considered, the overall computational complexity of the MAP-EM to estimate an $M \times N$ matrix is dominated by $\frac{1}{3 C^3}LMN^3$, with $L$ the number of iterations (typically $<10$) and $1/C$ the available data ratio, with $C$ typically large ($\approx 25$ for the standard movie ratings data). The algorithm is thus very fast. As each row $\bbf_i$ is treated as a signal and the signals can be estimated in sequence, the memory requirement is dominated by $N^2$ (to store the covariance matrix $\Sigma$).

\section{Numerical experiments}
\label{sec:numeric}

\subsection{Experimental Protocols}
The experimental protocols strictly follow those described in the literature~\cite{decoste2006collaborative, lawrence2009non, mackey2010, marlin2004collaborative, park2007applying, rennie2005fast, zhou10nonparametric}. 
The proposed method is evaluated on two movie rating prediction benchmark datasets, the \textit{EachMovie} dataset and the \textit{1M MovieLens} dataset.~\footnote{http://www.grouplens.org/} Two test protocols, the so-called ``weak generalization'' and ``strong generalization,''~\cite{marlin2004collaborative} are applied. 
\begin{itemize}
\item \textbf{Weak generalization} measures the ability of a method to generalize to other items rated by the \textit{same} users used for training the method. One known rating is randomly held out from each user's rating set to form a test set, the rest known ratings form a training set. The method is trained using the data in the training set, and its performance is evaluated over the test set. 
\item \textbf{Strong generalization} measures the ability of the method to generalize to some items rated by \textit{novel} users that have \textit{not} been used for training. The set of users is first divided into training users and test users. Learning is performed with all available ratings from the training set. To test the method, the ratings of each test users are further split into an observed set and a held-out set. The method is shown the observed ratings, and is evaluated by predicting the held-out ratings. 
\end{itemize}

The EachMovie dataset contains 2.8 million ratings in the range $\{1, \ldots, 6\}$ for 1,648 movies (columns) and 74,424 users (rows). Following the standard procedure~\cite{marlin2004collaborative}, users with fewer than 20 ratings and movies with less than 2 ratings are removed. This leaves us 36,656 users, 1,621 movies, and 2.5 million ratings (available data ratio $4.3\%$). We randomly select 30,000 users for the weak generalization, and 5,000 users for the strong generalization. The 1M MovieLens dataset contains 1 million ratings in the range $\{1, \ldots, 5\}$ for 3,900 movies (columns) and 6,040 users (rows). The same filtering leaves us 6,040 users, 3,592 movies, and 1 million ratings (available data ratio $4.6\%$).  We  randomly select 5,000 users for the weak generalization, and 1,000 users for the strong generalization. Each experiment is run 3 times and the average reported. 

The performance of the method is measured by the standard Normalized Mean Absolute Error (NMAE), computed by normalizing the mean absolute error by a factor for which random guessing produces a score of 1. The factor is 1.944 for EachMovie, and 1.6 for MovieLens. 

\subsection{Proposed Method Setup}
In contrast to most exiting algorithms in the literature, the proposed method, thanks to its simplicity, enjoys the advantage of having very few intuitive parameters. The covariance regularization parameter $\epsilon$ in~\eqref{eqn:cov:regularization} is set equal to $0.3$ (whose square root is one order of magnitude smaller than the maximum rating), the results being insensitive to this value as shown by the experiments. The number of iterations of the MAP-EM algorithm is fixed at $L=10$, beyond which the convergence of the algorithm is always observed. The noise $\bw_i$ in~\eqref{eqn:problem:vector} is neglected, i.e., $\Sigma_\bw$ is set to $\mathbf{0}$, as the movie rating datasets mainly involve missing data, the noise being implicit and assumed negligible. 

The experiments show that considering row $\bbf_i \in \mathbb{R}^N$ of the matrix $\bF \in \mathbb{R}^{M \times N}$ as signals  leads to slightly better results than taking columns or 2D sub-matrices. This means that each user is a signal, whose dimension is the number of movies.

As in previous works~\cite{zhou10nonparametric}, a post-processing that projects the estimated rating to an integer within the rating range is performed. 

A Matlab code of the proposed algorithm is available at \href{http://www.cmap.polytechnique.fr/~yu/research/MC/demo.html}{\it{http://www.cmap.polytechnique.fr/$\sim$yu/research/MC/demo.html}}.

\subsection{Results}

The results of the proposed method are compared with the best published ones including User Rating Profile (URP)~\cite{marlin2004collaborative}, Attitude~\cite{marlin2004modeling}, Maximum Margin Matrix Factorization (MMMF)~\cite{rennie2005fast}, Ensemble of MMMF (E-MMMF)~\cite{decoste2006collaborative}, Item Proximity based Collaborative Filtering (IPCF)~\cite{park2007applying}, Gaussian Process Latent Variable Models (GPLVM)~\cite{lawrence2009non}, Mixed Membership Matrix Factorization (M$^3$F)~\cite{mackey2010}, and Nonparametric Bayesian Matrix Completion (NBMC)~\cite{zhou10nonparametric}. For each of these methods, more than one results produced with different configurations are often reported, among which we systematically cite the best one. All these methods are significantly more complex than the one here proposed.

Tables~\ref{tab:each:movie} and~\ref{tab:1M:movielens} presents the results of various methods for both weak and strong generalizations on the two datasets. NBMC generates most often the best results, followed closely by the proposed method referred to as GM (Gaussian Model) and GPLVM, all of them outperforming the other methods. The results produced by the proposed GM, with a by far simpler model and faster algorithm, is in the same ballpark as those of NBMC and GPLVM, the difference with NMAE being smaller than about 0.005, marginal in the rating range that goes from 1 to 5 or 6. 

\begin{table}[htb]
\begin{center}
\begin{tabular}{|c||c|c|}
\hline
\textit{EachMovie}& Weak NMAE & Strong NMAE \\
\hline
\hline
URP~\cite{marlin2004collaborative} & 0.4422 & 0.4557 \\
\hline
Attitude~\cite{marlin2004modeling} & 0.4520 & 0.4550 \\
\hline 
MMMF~\cite{rennie2005fast} & 0.4397 & 0.4341\\
\hline
IPCF~\cite{park2007applying} & 0.4382 & 0.4365 \\
\hline
E-MMMF~\cite{decoste2006collaborative} & 0.4287 & 0.4301 \\
\hline
GPLVM~\cite{lawrence2009non} & 0.4179 & 0.4134 \\
\hline
M$^3$F~\cite{mackey2010} & 0.4293 & n/a \\
\hline
NBMC~\cite{zhou10nonparametric} & \textbf{0.4109} & \textbf{0.4091} \\
\hline 
GM (proposed) & 0.4164 & 0.4163 \\
\hline
\end{tabular}
\end{center}
\caption{NMAEs generated by different methods for EachMovie database. The smallest NMAE is in boldface.} 
\label{tab:each:movie}
\end{table}

\begin{table}[htb]
\begin{center}
\begin{tabular}{|c||c|c|}
\hline
\textit{1M MovieLens} & Weak NMAE & Strong NMAE \\
\hline
\hline
URP~\cite{marlin2004collaborative} & 0.4341 & 0.4444 \\
\hline
Attitude~\cite{marlin2004modeling} & 0.4320 & 0.4375 \\
\hline 
MMMF~\cite{rennie2005fast} & 0.4156 & 0.4203 \\
\hline
IPCF~\cite{park2007applying} & 0.4096 & 0.4113 \\
\hline
E-MMMF~\cite{decoste2006collaborative} & 0.4029 & 0.4071 \\
\hline
GPLVM~\cite{lawrence2009non} & 0.4026 & 0.3994 \\
\hline
M$^3$F~\cite{mackey2010} & n/a & n/a \\
\hline
NBMC~\cite{zhou10nonparametric} & \textbf{0.3916} & 0.3992 \\
\hline 
GM (proposed) & 0.3959 & \textbf{0.3928} \\
\hline
\end{tabular}
\end{center}
\caption{NMAEs generated by different methods for 1M MovieLens database. The smallest NMAE is in boldface.} 
\label{tab:1M:movielens}
\end{table}

\vspace{-4ex}
\section{Conclusion and Discussion}
\label{sec:conclusion}
We have shown that a Gaussian model and a MAP-EM algorithm provide a simple and computational efficient solution for matrix completion, leading to results in the same ballpark as state-of-the-art ones for movie rating prediction. 

Future work may go in several directions. On the one hand, the proposed conceptually simple and computationally efficient method may provide a good baseline for further refinement, for example by incorporating user and item bias or meta information~\cite{zhou10nonparametric}. On the other hand, Gaussian mixture models (GMM) that have been shown to bring dramatic improvements over single Gaussian models in image inpainting~\cite{yu2010PLE}, are expected to better capture different characteristics of various categories of movies (comedy, action, etc.) and classes of users (age, gender, etc.). However, no significant improvement by GMM over Gaussian model has yet been observed for movie rating prediction. This needs to be further investigated, and such improvement might come from proper grouping and initialization.

{\small
\vspace{2ex} \noindent
{\bf Acknowledgments:} Work partially supported by NSF, ONR, NGA, ARO, and NSSEFF. We thank S. Mallat for co-developing the proposed framework, and M. Zhou and L. Carin for their comments and help with the data.
}

\vspace{-3ex}

\bibliographystyle{plain}

\end{document}